\documentclass[letterpaper, 10 pt, journal, twoside]{IEEEtran}
\usepackage{graphicx}      
\usepackage{cite}
\usepackage{array}
\usepackage{rotating}
\usepackage{tabularray}
\usepackage{graphics} 
\usepackage{epsfig}   
\usepackage[caption=false]{subfig}
\usepackage{mathptmx} 
\usepackage{times}    
\usepackage{amsmath}  %
\usepackage{amssymb}  %
\usepackage[usenames,dvipsnames]{xcolor}
\usepackage{booktabs}
\usepackage[hidelinks]{hyperref}
\usepackage[none]{hyphenat}
\usepackage[short]{optidef}
\usepackage{cases}
\usepackage{bm}
\usepackage{multirow}
\usepackage{tabularx}

\usepackage{algorithm}
\usepackage[
  noEnd          = true, 
  italicComments = true,  
  rightComments  = true,  
  commentColor   = black, 
  spaceRequire   = true   
]{algpseudocodex}

\makeatletter
\renewcommand{\ALG@beginalgorithmic}{\small}

\makeatother

\newcommand{\IfAlg}{\ensuremath{\textbf{if}}}
\newcommand{\ElseAlg}{\ensuremath{\textbf{else}}}

\newcommand{\ContinueAlg}{\ensuremath{\textbf{continue}}}

\newcommand{\ThenAlg}{\ensuremath{\textbf{then}}}
  
\usepackage[normalem]{ulem}
\algnewcommand{\IIf}[1]{\State\algorithmicif\ #1\ \algorithmicthen}
\algnewcommand{\EndIIf}{\unskip\ \algorithmicend\ \algorithmicif}

\usepackage{tikz}
\usetikzlibrary{shapes, arrows.meta, positioning,calc}
\usepackage{pgfplots}
\pgfplotsset{compat=1.18}        

\usepackage{multirow}

\usepackage{acronym}
\newacro{DSM}{Direct Stiffness Method}
\newacro{FE}{Finite Element}
\newacro{FEM}{Finite Element Method}
\newacro{LU}{Lower-Upper}
\newacro{MB}{Multi-Body}
\newacro{ODE}{Ordinary Differential Equation}
\newacro{DAE}{Differential-Algebraic Equation}
\newacro{BVP}{Boundary Value Problem}
\newacro{ARD}{Artificial Race Driver}
\newacro{OCP}{Optimal Control Problem}
\newacro{MPC}{Model Predictive Control}
\newacro{EMPC}{Economic Model Predictive Control}
\newacro{MP}{Motion Primitive}
\newacro{RRT}{Rapidly-exploring Random Tree}
\newacro{RRT*}{Rapidly-exploring Random Tree Star}
\newacro{MPTree}{Motion Primitive Tree}
\newacro{V2X}{Vehicle-to-Everything}
\newacro{VRU}{Vulnerable Road User}
\newacro{SIL}{Software-in-the-Loop}
\newacro{HIL}{Hardware-in-the-Loop}
\newacro{VIL}{Vehicle-in-the-Loop}
\newacro{ADAS}{Advanced Driver Assistance System}
\newacro{AD}{Autonomous Driving}
\newacro{CAV}{Connected and Automated Vehicle}
\newacro{CC}{Cruise Control}
\newacro{ACC}{Adaptive Cruise Control}
\newacro{CACC}{Cooperative Adaptive Cruise Control}
\newacro{V2V}{Vehicle-to-Vehicle}
\newacro{V2I}{Vehicle-to-Infrastructure}
\newacro{V2N}{Vehicle-to-Network}
\newacro{MCTS}{Monte Carlo Tree Search}
\newacro{QP}{Quadratic Programming}
\newacro{PMP}{Pontryagin's Maximum Principle}
\newacro{MLT}{Minimum Lap Time}
\newacro{FBGA}{Forward-Backward method with Generic Acceleration constraints}
\newacro{QSS}{Quasi-Steady-State}
\newacro{FB}{Forward-Backward}

\definecolor{BlueMATLAB}{rgb}{0.00000,0.44700,0.74100}%
\definecolor{RedMATLAB}{rgb}{0.85000,0.32500,0.09800}%
\definecolor{BrownMATLAB}{rgb}{0.92900,0.69400,0.12500}%
\definecolor{PurpleMATLAB}{rgb}{0.49400,0.18400,0.55600}%
\definecolor{GreenMATLAB}{rgb}{0.46600,0.67400,0.18800}%
\definecolor{LigthBlueMATLAB}{rgb}{0.30100,0.74500,0.93300}%
\definecolor{DarkRedMATLAB}{rgb}{0.63500,0.07800,0.18400}%
\definecolor{MATLAB1}{rgb}{0.00000,0.44700,0.74100}%
\definecolor{MATLAB2}{rgb}{0.85000,0.32500,0.09800}%
\definecolor{MATLAB3}{rgb}{0.92900,0.69400,0.12500}%
\definecolor{MATLAB4}{rgb}{0.49400,0.18400,0.55600}%
\definecolor{MATLAB5}{rgb}{0.46600,0.67400,0.18800}%
\definecolor{MATLAB6}{rgb}{0.30100,0.74500,0.93300}%
\definecolor{MATLAB7}{rgb}{0.63500,0.07800,0.18400}%
\newcommand{\eg}{\emph{e.g.},}
\newcommand{\ie}{\emph{i.e.},}
\usepackage{soul}

\usepackage{subfig}

\newif\ifarxiv
  \arxivtrue

\newcommand\copyrighttext{%
  \footnotesize \textcopyright 2025 IEEE. Personal use of this material is permitted.
  Permission from IEEE must be obtained for all other uses, in any current or future
  media, including reprinting/republishing this material for advertising or promotional
  purposes, creating new collective works, for resale or redistribution to servers or
  lists, or reuse of any copyrighted component of this work in other works.}
\newcommand\copyrightnotice{%
\begin{tikzpicture}[remember picture,overlay]
\node[anchor=north,yshift=-10pt] at (current page.north) 
  {\fbox{\parbox{\dimexpr\textwidth-\fboxsep-\fboxrule\relax}{\copyrighttext}}};
\end{tikzpicture}%
}

\newcommand\citeRAL{%
     Please refer to the published version of this article:\\
      \href{https://doi.org/10.1109/LRA.2025.3643297}{M. Piazza, M. Piccinini, S. Taddei, F. Biral, and E. Bertolazzi,
      "Real-Time Velocity Profile Optimization for Time-Optimal Maneuvering With Generic Acceleration Constraints,"
      \textit{IEEE Robotics and Automation Letters}, vol. 11, no. 2, pp. 1674-1681, 2026.}
      \textsc{doi:} \href{https://doi.org/10.1109/LRA.2025.3643297}{10.1109/LRA.2025.3643297}
 }
\newcommand\RALnotice{%
\begin{tikzpicture}[remember picture,overlay]
\node[anchor=north,yshift=-50pt] at (current page.north) 
  {\fbox{\parbox{\dimexpr\textwidth-\fboxsep-\fboxrule\relax}{\citeRAL}}};
\end{tikzpicture}%
}

\title{\LARGE \bf Real-time Velocity Profile Optimization for Time-Optimal Maneuvering with Generic Acceleration Constraints} 
\author{Mattia Piazza$^{1}$, Mattia Piccinini$^{2}$, Sebastiano Taddei$^{1,3}$, Francesco Biral$^{1}$, and Enrico Bertolazzi$^{1}$%
\thanks{Manuscript received: July, 11$^\mathrm{th}$, 2025; Revised September, 19$^\mathrm{th}$, 2025; Accepted November, 19$^\mathrm{th}$, 2025.}
\thanks{This paper was recommended for publication by
Editor Lucia Pallottino upon evaluation of the Associate Editor and Reviewers' comments.} 
\thanks{$^{1}$ Department of Industrial Engineering,
University of Trento, 38123 Trento, Italy, {\tt\small name.surname@unitn.it}.}%
\thanks{$^{2}$ Professorship of Autonomous Vehicle Systems, Technical University
of Munich, 85748 Garching, Germany; Munich Institute of Robotics and
Machine Intelligence (MIRMI), {\tt\small mattia.piccinini@tum.de}.}
\thanks{$^{3}$Department of Electrical and Information Engineering, Politecnico di Bari, 70125 Bari, Italy, {\tt\small s.taddei@phd.poliba.it}.}
\thanks{This work was partly supported by the European Union - Next Generation
EU - under the National Recovery and Resilience Plan (NRRP), Mission
4 Component 1 Investment 3.4 - Decree No. 351 of Italian Ministry of University and Research - Concession Decree No. 2152 of the Italian Ministry of University and Research, Project code D93C22000500001, within the
Italian National Program PhD Programme in Autonomous Systems (DAuSy).}
\thanks{Digital Object Identifier (DOI): see top of this page.}
}

\begin{document}
\copyrightnotice
\RALnotice
\newpage
\maketitle
%
%
%
%
\begin{abstract}
The computation of time-optimal velocity profiles along prescribed paths, subject to generic acceleration constraints, is a crucial problem in robot trajectory planning, with particular relevance to autonomous racing. However, the existing methods either support arbitrary acceleration constraints at high computational cost or use conservative box constraints for computational efficiency. We propose FBGA, a new \underline{F}orward-\underline{B}ackward algorithm with \underline{G}eneric \underline{A}cceleration constraints, which achieves both high accuracy and low computation time. FBGA operates forward and backward passes to maximize the velocity profile in short, discretized path segments, while satisfying user-defined performance limits. Tested on five racetracks and two vehicle classes, FBGA handles complex, non-convex acceleration constraints with custom formulations. Its maneuvers and lap times closely match optimal control baselines (within $0.11\%$-$0.36\%$), while being up to three orders of magnitude faster. FBGA maintains high accuracy even with coarse discretization, making it well-suited for online multi-query trajectory planning.
Our open-source \texttt{C++} implementation is available at: \url{https://github.com/DRIVEWISE/FBGA}.
\end{abstract}
%
%
%
\section{INTRODUCTION} \label{sec:intro}
\begin{figure}[t]
  \centering
  \subfloat[Path, velocity ($v_x$) and longitudinal acceleration ($a_x$).]{\includegraphics[width=0.98\columnwidth]{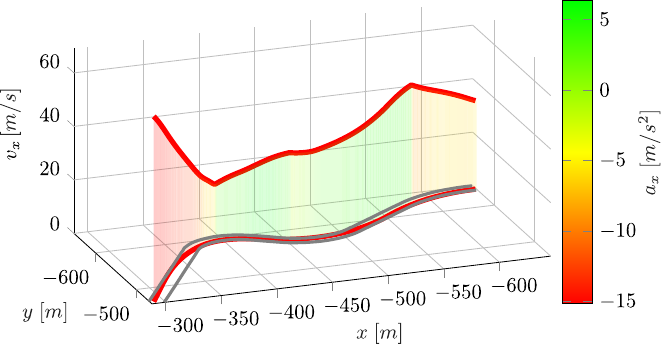}\label{fig:3DTrajB}}\\
  \subfloat[Lateral acceleration ($a_y$) and curvature ($\kappa$), with their local maxima.]{\includegraphics[width=0.98\columnwidth]{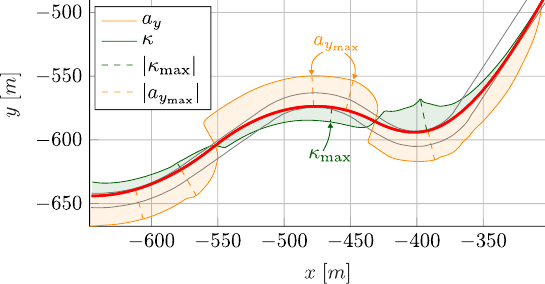}\label{fig:3DTrajA}}
  \caption{(a) Time-optimal velocity and longitudinal acceleration profiles computed by our FBGA along the first two corners of the Catalunya circuit. (b) Lateral acceleration and curvature profiles. Unlike QSS \cite{Brayshaw2005}, FBGA does not assume that peak lateral accelerations ($a_{y_{\max}}$) occur at curvature peaks ($\kappa_{\max}$): in the second corner,  $\kappa_{\max}$ happens between two $a_{y_{\max}}$ peaks.
  }
  \vspace{-0.1cm}
  \label{fig:3DTraj}
\end{figure}

Time-optimal velocity planning under acceleration constraints is a key problem in robotics, with applications in autonomous racing (Fig. \ref{fig:3DTraj}) \cite{Ogretmen2024,frego2017semi,Piccinini2024}, drone flight \cite{Romero2022}, manipulators \cite{Bobrow1985}, and mobile robot navigation \cite{Loknar2023,Consolini2022}. These applications use point-mass models to compute time-optimal velocity profiles along fixed paths, subject to acceleration limits. Fast online generation of such profiles is essential in dynamic environments, where trajectory planners must evaluate many candidate maneuvers, often with graph- or sampling-based frameworks \cite{Ogretmen2024,piazza2024mptree,Romero2022,trauth2024frenetix}.

However, the existing methods for minimum-time velocity planning face a trade-off: \acp{OCP} \cite{piccinini2025howoptimal} and \ac{QSS} \cite{Veneri2020,rodegher2024effect} offer high accuracy, but their high computational cost does not allow online multi-query planning. Conversely, semi-analytical \ac{FB} and sequential methods \cite{frego2017semi,Consolini2022} are faster but limited to box-shaped acceleration bounds, which are overly conservative for racing vehicles. Accurate performance modeling of racing vehicles requires complex g-g-v diagrams (Fig. \ref{fig:ggvA}) \cite{limebeer2024asymmetric,Veneri2020,piccinini2024impacts, werner2025quasi}, describing the speed-dependent coupling of the lateral and longitudinal acceleration limits.

This paper introduces a real-time \ac{FBGA}, matching the accuracy of \acp{OCP} while being up to three orders of magnitude faster, making it a suitable building block for online sampling-based trajectory planners.
\subsection{Related Work} \label{sec:relatedwork}
In robotics and autonomous racing, the methods to compute minimum-time velocity profiles on a given path fall into three families: (1) \ac{OCP} with direct and indirect methods, (2) \ac{QSS}, and (3) \ac{FB} or related semi-analytical strategies. 
\subsubsection{Optimal Control Methods} \label{sec:OCPMethods}
Minimum-time \acp{OCP} are typically solved using direct \cite{Rowold2023} or indirect \cite{biral2016notes} methods to jointly optimize path and velocity. They are used in offline lap-time minimization with multibody models \cite{Lot2016}, and in online planning with point-mass models for autonomous racing \cite{Piccinini2025_ICRA,Rowold2023,piccinini2025howoptimal}. However, even with simplified models, \acp{OCP} remain computationally intensive and are not suitable for online evaluation of multiple trajectories in dynamic, non-convex race scenarios\cite{Piccinini2024}.
\subsubsection{Quasi-Steady-State Methods} \label{sec:QSSMethods} 
QSS methods \cite{Brayshaw2005} solve the same problem addressed in this paper: minimum-time velocity planning along a fixed path under g-g-v acceleration constraints. QSS splits the path at the curvature apexes and solves a sequence of nonlinear programs (NLPs) to compute the optimal velocity profiles. While effective, QSS methods are computationally expensive and unsuitable for online planning. For instance, \cite{Biniewicz2024} applied QSS to a point-mass motorcycle model with jerk bounds for offline lap-time optimization, and \cite{Veneri2020,rodegher2024effect} used QSS in free-trajectory planning with point-mass car models, yet their computational times were too high for online planning. Moreover, QSS critically depends on apex selection for numerical convergence, and assumes that the maximum lateral acceleration is at the path apexes, which is not always the time-optimal maneuver (Fig. \ref{fig:3DTraj}).
\subsubsection{Forward-Backward, Semi-Analytical and Alternative Methods} \label{sec:FBMethods}
In race car trajectory optimization, two-step methods combining forward-backward (FB) velocity planning and minimum-curvature path generation were proposed in \cite{kapania2016sequential,heilmeier2020minimum}, while \cite{lenzo2020simple} applied a similar iterative approach along a fixed path. However, their methods required tenths of seconds to minutes, and could not allow online re-planning. Conversely, the semi-analytical FB algorithm in \cite{frego2017semi} enabled online time-optimal speed planning, yet with box-constrained accelerations (rectangular g-g diagram). Similar speed planning problems were solved with range reduction \cite{Cabassi2018}, sequential line search \cite{Consolini2022}, and convexification \cite{Consolini2024}, adding acceleration rate and jerk limits. However, the box acceleration constraints used by \cite{frego2017semi,Cabassi2018,Consolini2022,Consolini2024} remained the main limiting factor, being very conservative for racing vehicles, whose performance envelope is typically defined by speed-dependent, diamond-shaped g-g-v diagrams \cite{piccinini2024impacts,Rowold2023,langmann2025online}.
Still, such simplified constraints were used in sampling-based planners for autonomous racing with cars \cite{piazza2024mptree} and drones \cite{Romero2022}, where computational efficiency is prioritized over accurate performance modeling. We aim to extend these methods to support generic acceleration constraints while maintaining computational efficiency.
\subsubsection{Critical Summary}
To the best of our knowledge, existing methods are limited by at least one of the following:
\begin{enumerate}
  \item \ac{OCP} and \ac{QSS} methods are not suitable for online multi-query planning due to their relatively high computational cost \cite{Veneri2020,Biniewicz2024,piccinini2025howoptimal}. Also, QSS assumes that the maximum lateral acceleration is reached at the corner apexes \cite{Brayshaw2005}, which may be non-time-optimal.
  \item Semi-analytical \ac{FB} \cite{frego2017semi}, sequential and convexified methods \cite{Cabassi2018,Consolini2022,Consolini2024} 
  are computationally efficient for online speed planning but limited to rectangular g-g acceleration constraints, which poorly represent the true performance envelope of vehicles and robots. 
\end{enumerate}
\begin{figure}[]
  \centering
  \subfloat[]{\includegraphics[width=0.495\columnwidth]{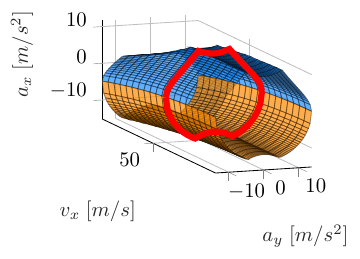}\label{fig:ggvA}}
  \subfloat[]{\includegraphics[width=0.495\columnwidth]{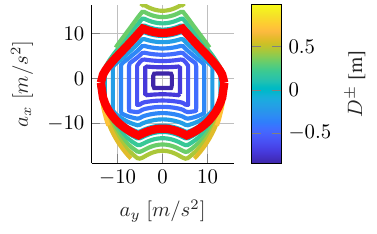}\label{fig:ggvB}}
  \caption{(a) g-g-v diagram of the racing motorcycle model \cite{Biral2009} used in Sec.~\ref{sec:results}. (b) Output of the $D^\pm$ signed distance function (Algorithm~\ref{alg_signed_distance}) for the red slice in (a). $D^\pm$ ranges from $-1$ (inside the g-g-v envelope) to $+\infty$ (outside). The red line denotes the envelope boundary, where $D^\pm = 0$.
  }
  \label{fig:generic_GG_nc_Dpm}
  \vspace{-0.1cm}
\end{figure}
\subsection{Contributions}
To address the limitations of existing methods, we propose the following contributions:
\begin{enumerate}
  \item FBGA: a new method for time-optimal velocity planning along a given path. For the first time, FBGA enables real-time planning with generically-shaped acceleration constraints (g-g-v diagrams).
  \item We validate FBGA on different racetracks, using both race car and motorcycle examples, and show that it can accurately compute time-optimal velocity profiles for different vehicle classes and acceleration constraints.
  \item We compare our approach against \acp{OCP}, achieving lower computational cost (up to three orders of magnitude) with the same level of accuracy, showing the potential for online planning in autonomous racing.
\end{enumerate}
\section{METHODOLOGY}
\subsection{Time-Optimal Velocity Planning Problem}
We consider the following velocity planning problem along a given path:
\begin{mini!}
	{\mathclap{a_x}}
	{T\label{eq_OCP_cost}}
	{\label{eq_OCP_problem}}{}
	\addConstraint{}{\dot{s}=v_x, \hspace{0.2cm} \dot{v}_x=a_x \label{eq_OCP_dyna}}
    \addConstraint{}{
    \Gamma_y^-(v_x) \, \leq \, a_y = \kappa(s) \, v_x^2 \, \leq \, \Gamma_y^+(v_x) \label{eq_OCP_ay_constr}
    }
    \addConstraint{}{
    \Gamma_x^-\big( a_y,v_x \big) \, \leq \, a_x \, \leq \, \Gamma_x^+\big( a_y,v_x \big)
    \label{eq_OCP_ax_constr}
    }
    \addConstraint{}{v_x(0) = v_{\mathrm{ini}}, \; s(0) = 0, \; s(T) = L. \label{eq_OCP_bcs}}
\end{mini!}
Where $T$ is the maneuver time to be minimized. The system dynamics \eqref{eq_OCP_dyna} is a double integrator, where the states $\{s,v_x\}$ are the curvilinear abscissa along the path and the longitudinal velocity, while the control input $a_x$ is the longitudinal acceleration. The control $a_x$ is bounded by the acceleration constraints \eqref{eq_OCP_ay_constr}-\eqref{eq_OCP_ax_constr}, where $\kappa(s)$ is the curvature of the path as a function of $s$, $a_y = \kappa(s) \, v_x^2$ is the lateral acceleration, and $\Gamma_x^\pm(\cdot)$ and $\Gamma_y^\pm(\cdot)$ are functions providing the lower and upper bounds of the lateral and longitudinal accelerations. Finally, the boundary conditions \eqref{eq_OCP_bcs} specify the initial velocity $v_{\mathrm{ini}}$ and the final travelled distance $L$. In \eqref{eq_OCP_problem}, all the variables are time-dependent, and the notation $\dot{\mathrm{x}}$ denotes the time derivative of the variable $\mathrm{x}$.

Assuming the lateral acceleration $a_y = \kappa(s) \, v_x^2$ places the model in the quasi-steady-state class, along a given path.
Our formulation \eqref{eq_OCP_problem} condenses all dynamic nonlinearities in the acceleration constraints \eqref{eq_OCP_ay_constr}-\eqref{eq_OCP_ax_constr}, which, for vehicles, define the g-g-v diagram. This diagram captures key physical effects, including tire saturation, actuation limits, aerodynamic drag, load transfer, and front/rear wheel lift in motorcycles, and expresses the maximum performance envelope in terms of center-of-mass accelerations.

Variants of the problem \eqref{eq_OCP_problem} were solved with optimal control for vehicle motion planning \cite{Novi2020,piccinini2025howoptimal}, but their high computational cost hinders online planning of multiple trajectories in dynamic scenarios. This paper introduces FBGA, a novel algorithm that solves \eqref{eq_OCP_problem} with high accuracy and significantly lower computation time.
\subsubsection{Solution with Acceleration Constraints at the Borders}
Let us start by considering the problem \eqref{eq_OCP_problem} with the acceleration constraints \eqref{eq_OCP_ay_constr}-\eqref{eq_OCP_ax_constr} enforced only at borders ($s=0$ and $s=L$). Since the maneuver time $T$ in \eqref{eq_OCP_cost} is unknown, to solve the problem we perform a change of independent variable, using the curvilinear abscissa $s$ instead of the time $t$. This yields:
\begin{mini!}
	{\mathclap{a_x}}{\int_0^L\dfrac{\mathrm{d} s}{v_x(s)}
  \label{eq_OCP_cost_s}}
	{\label{eq_OCP_problem_s}}{}
	\addConstraint{}{v_x^\prime(s)\,v_x(s) = a_x(s) \label{eq_OCP_dyna_s}}
  \addConstraint{}{v_x(0) = v_{\mathrm{ini}} \,, \; \text{with \eqref{eq_OCP_ay_constr}-\eqref{eq_OCP_ax_constr} at} ~ s=0 ~ \text{and} ~ s=L
  \label{eq_OCP_bcs_s}}
\end{mini!}
where $v_x^\prime(s) = \mathrm{d} v_x(s)/\mathrm{d} s$. Using the Pontryagin's Maximum
Principle, the problem \eqref{eq_OCP_problem_s} yields an analytical solution, which simplifies considering a constant longitudinal acceleration:
\begin{equation}
v_x(s) = \left.\sqrt{2 \, \int_0^s a_x(\zeta) \, \mathrm{d}\zeta + v_{\mathrm{ini}}^2}\,\right|_{a_x(\zeta)=a_x} = \sqrt{2 \, s \, a \, + \, v_{\mathrm{ini}}^2} \; .
\label{eq_vx_s_no_constr}
\end{equation}
Furthermore, the resulting maneuver time is: 
\begin{equation}
  T = \displaystyle \int_0^L \frac{\mathrm{d} s}{v_x(s)} = \frac{-v_{\mathrm{ini}} + \sqrt{2 \, a \, L + v_{\mathrm{ini}}^2}}{a} \, .
  \label{eq_OCP_T_s}
\end{equation}
Sec. \ref{sec:FB_algorithm} will use \eqref{eq_vx_s_no_constr}-\eqref{eq_OCP_T_s} on short discretized segments of the path, where $a_x$ will be assumed piecewise constant and on the border of the acceleration constraints \eqref{eq_OCP_ay_constr}-\eqref{eq_OCP_ax_constr}.\footnote{Assuming a constant longitudinal acceleration within short path segments is consistent with direct optimal control methods, where the control inputs are typically held constant in discretized mesh intervals.}
This assumption introduces discontinuities in $a_x$ at the segment boundaries, which is accepted for real-time trajectory planning, but is not representative of the real vehicle dynamics. Jerk constraints could be enforced by increasing the algorithm's complexity, and will be part of future work.
\subsection{Signed Distance from the g-g-v Envelope} \label{sec:signed_distance}
Before describing the FBGA algorithm, we need to define an auxiliary function that computes the signed distance of a point from the g-g-v acceleration envelope. The g-g-v is defined by the functions $\Gamma_y^\pm(v_x)$ and $\Gamma_x^\pm(a_y,v_x)$ in \eqref{eq_OCP_ay_constr}-\eqref{eq_OCP_ax_constr}, which represent the bounds of the lateral and longitudinal accelerations. An example of g-g-v shape is given in Fig. \ref{fig:ggvA}, where a slice at constant speed is shown in red.
Algorithm \ref{alg_signed_distance} introduces a new function, named $D^\pm(\cdot)$, to compute the signed distance of a point $P = (a_x,a_y,v_x)$ from the g-g-v. As shown in Fig. \ref{fig:generic_GG_nc_Dpm}, $D^\pm(P)$ yields a positive value if $P$ is outside the g-g-v envelope, and a negative value otherwise. The function is exactly zero if $P$ is along the envelope.

\begin{algorithm}[]
  \caption{$\textsc{D}^\pm$ signed distance function}
  \label{alg_signed_distance}
  \begin{algorithmic}[1]
    \State \textbf{Input:} point $P = (a_x,a_y,v_x)$ 
    \State \textbf{Output:} signed distance $D$ of $P$ from the g-g-v envelope
    \State $a_{y_{\mathrm{clip}}}$ $\gets$ $\Call{clip}{a_y, \Gamma_{y}^-(v_x), \Gamma_y^+(v_x)}$\,; \label{alg:dpm_clip}
    \State $a_{x_{\min}}$ $\gets$ $\Gamma_x^-(a_{y_{\mathrm{clip}}},v_x)$\,; \; $a_{x_{\max}}$ $\gets$ $\Gamma_x^+(a_{y_{\mathrm{clip}}},v_x)$\,; \label{alg:dpm_ax_bounds}
    \State $D$ $\gets$ $\Lambda\Big(\,\Phi\big(a_x, a_{x_{\min}}, a_{x_{\max}}\big),\, \Phi\big(a_y, \Gamma_{y}^-(v_x), \Gamma_{y}^+(v_x)\big)\,\Big)$\,; \label{alg:dpm_lambda_phi}
  \end{algorithmic}
\end{algorithm}
Algorithm \ref{alg_signed_distance} starts by clipping the lateral acceleration $a_y$ of $P$ to the g-g-v bounds (line~\ref{alg:dpm_clip}). 
Then, it retrieves the longitudinal acceleration limits $\{a_{x_{\min}}, a_{x_{\max}}\}$ for the clipped $a_y$ and the given $v_x$ (line~\ref{alg:dpm_ax_bounds}). Finally, it computes the signed distance $D$ using the functions $\Lambda(\cdot)$ and $\Phi(\cdot)$ (line~\ref{alg:dpm_lambda_phi}). $\Lambda(\cdot)$ has the shape of a pyramid pointing downwards, forming a square trace on the horizontal plane:
\begin{equation}
  \Lambda(x,\,y) \, = \, \max\big( x-1, \, -1-x, \, y-1, \, -1-y \big)
  \label{eq:pyramid}
\end{equation}
The function $\Phi(\cdot)$ maps a given range $[\mathrm{x}_{\min}, \mathrm{x}_{\max}]$ to $[-1, +\infty)$, enabling a consistent evaluation of the $\Lambda(\cdot)$ function: 
\begin{equation}
  \Phi(\mathrm{x}, \, \mathrm{x}_{\min}, \, \mathrm{x}_{\max}) \, = \, 2 \,\, \frac{\mathrm{x}-\mathrm{x}_{\min}}{\mathrm{x}_{\max}-\mathrm{x}_{\min}}-1
\end{equation}
The resulting function $D^\pm(\cdot)$ is plotted in Fig. \ref{fig:ggvB}.
\subsection{FBGA Algorithm} \label{sec:FB_algorithm}
\begin{figure*}
  \centering
  \includegraphics[width=0.99\textwidth]{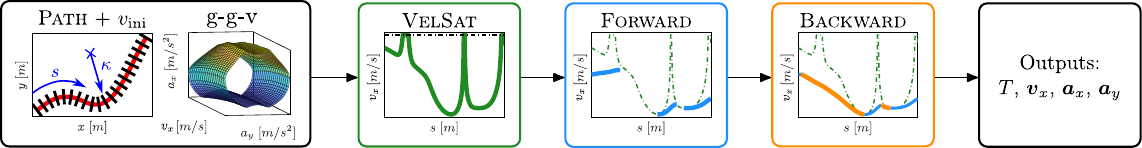}
  \vspace{-0.1cm}
  \caption{Main phases of our FBGA, described in Algorithm \ref{alg_FWBW}. FBGA takes as input the path (vectors of curvilinear abscissas $\boldsymbol s$ and curvatures $\boldsymbol \kappa$), the initial speed $v_{\mathrm{ini}}$, and the g-g-v acceleration constraints. It outputs vectors with the speed and acceleration profiles $\{\boldsymbol{v}_x, \boldsymbol{a}_x, \boldsymbol{a}_y\}$, and the maneuver time $T$.}
  \label{fig:FB_diagram}
  \vspace{-0.1cm}
\end{figure*}
\begin{figure}[]
  \centering
  \includegraphics[width=0.98\columnwidth]{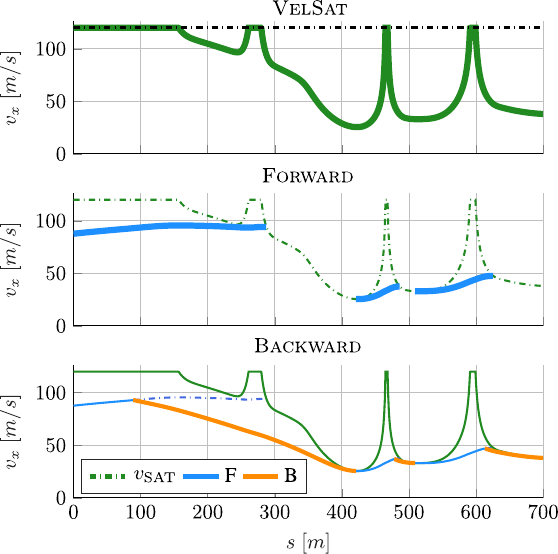}
  \vspace{-0.1cm}
  \caption{Phases of the FBGA method on an example scenario: maximum speed given the lateral acceleration bounds (top plot), forward pass (center plot, with blue lines when successful), and backward pass (bottom plot, with orange lines when editing invalid forward pass segments).
  }
  \label{fig:algorithm_phases}
  \vspace{-0.1cm}
\end{figure}
As depicted in Fig. \ref{fig:FB_diagram}, the proposed FBGA method solves the problem \eqref{eq_OCP_problem} using the following inputs:
\begin{enumerate}
  \item The path to be driven, in the form of vectors of curvilinear abscissas $\boldsymbol s = [s_1, \ldots, s_N]$ and corresponding curvatures $\boldsymbol \kappa = [\kappa_1, \ldots, \kappa_N] $.
  \item The initial speed $v_{\mathrm{ini}}$ and the top speed $v_{\max}$.
  \item The g-g-v acceleration constraints, expressed by the arbitrary function $\Gamma_y^\pm(v_x)$ and $\Gamma_x^\pm(a_y,v_x)$ in \eqref{eq_OCP_ay_constr}-\eqref{eq_OCP_ax_constr}, which can represent any shape of the g-g-v envelope.
\end{enumerate}
FBGA outputs arrays of speed and acceleration profiles $\{\boldsymbol{v}_x, \boldsymbol{a}_x, \boldsymbol{a}_y\}$\footnote{In post-processing, FBGA returns $v_x(t)$, $a_x(t)$, and $a_y(t)$ $\forall\, t \in [0,T]$.}, and the maneuver time $T$. Algorithm \ref{alg_FWBW} and Fig. \ref{fig:FB_diagram} provide an overview of the FBGA method. For each of the $N-1$ path segments, Algorithm \ref{alg_FWBW} performs the following three steps: calculation of the maximum saturated speed given the lateral acceleration bounds (line~\ref{alg:fb_velsat}), forward pass (line~\ref{alg:fb_forward}), and backward pass (line~\ref{alg:fb_backward}). The total maneuver time $T$ is obtained as the sum of the individual segment durations (line~\ref{alg:fb_time}), each computed via \eqref{eq_OCP_T_s}. The lateral acceleration vector $\boldsymbol{a}_y$ is finally given by the Hadamard product of the curvature $\boldsymbol{\kappa}$ and the squared velocity profile $\boldsymbol{v}_x^2$ (line~\ref{alg:fb_time}). 

Fig. \ref{fig:algorithm_phases} illustrates the three main phases of the FBGA on an example scenario.
These phases, detailed in Algorithms \ref{alg_Vmax}-\ref{alg_BW}, share a custom root-finding routine, named \textsc{SOLVE}, which computes function zeros with a custom non-derivative method. 
Optimality proofs of our FBGA with generic acceleration constraints are an open research question. However, optimality proofs exist for convex or rectangular g-g shapes and linear dynamics (in the control) \cite{velenis2008minimum,frego2017semi}.
\begin{algorithm}[]
  \caption{FBGA method}
  \label{alg_FWBW}
  \begin{algorithmic}[1]
    \State \textbf{Input:} $\boldsymbol s$, $\boldsymbol \kappa$, $v_{\mathrm{ini}}$, $v_{\max}$, $\Gamma_y^-$, $\Gamma_y^+$, $\Gamma_x^-$, $\Gamma_x^+$
    \State \textbf{Output:} $\boldsymbol{v}_x$, $\boldsymbol{a}_x$, $\boldsymbol{a}_y$, $T$
    \State $\boldsymbol{v}_{\mathrm{sat}} \gets$ \textsc{VelSat}($\boldsymbol s$, $\boldsymbol \kappa$, $v_{\max}$, $\Gamma_y^-$, $\Gamma_y^+$)\,;\label{alg:fb_velsat}
    \State $\boldsymbol{v}_{x}, \boldsymbol{a}_{x} \gets \textsc{F}(\boldsymbol s, \boldsymbol \kappa, v_{\mathrm{ini}}, \Gamma_y^-, \Gamma_y^+, \Gamma_x^-, \Gamma_x^+, \boldsymbol{v}_{\mathrm{sat}})$\,; \label{alg:fb_forward}
    \State $\boldsymbol{v}_{x}, \boldsymbol{a}_{x} \gets \textsc{B}(\boldsymbol s, \boldsymbol \kappa, \Gamma_y^-, \Gamma_y^+, \Gamma_x^-, \Gamma_x^+, \boldsymbol{v}_{\mathrm{sat}}, \boldsymbol{v}_{x}, \boldsymbol{a}_{x})$\,; \label{alg:fb_backward}
    \State $N \gets \mathrm{size}(\boldsymbol s)$\,; 
    \State $T \gets  \sum_{i=1}^{N-1} \frac{-\boldsymbol{v}_{x_i} + \sqrt{2 \, \boldsymbol{a}_{x_i} (\boldsymbol{s}_{i+1}-\boldsymbol{s}_{i}) \, + \, \boldsymbol{v}_{x_i}^2}}{\boldsymbol{a}_{x_i}}$\,; \hspace{0.04cm} $\boldsymbol{a}_y \gets \boldsymbol \kappa \odot \boldsymbol{v}_x^2$\,; \label{alg:fb_time}
  \end{algorithmic}
\end{algorithm}
\subsubsection{Maximum Speed given the Lateral Acceleration Bounds} \label{sec:VelSat}
As shown in the top plot of Fig. \ref{fig:algorithm_phases}, FBGA first computes a vector of saturated velocities $\boldsymbol{v}_{\mathrm{sat}}$, which are the maximum speeds to comply with the lateral acceleration bounds $\Gamma_y^\pm(v_x)$. This calculation is done by the \textsc{VelSat} function, provided in Algorithm \ref{alg_Vmax}. The function loops over the path points (line~\ref{alg:velsat_for}), and computes the corresponding saturated speed $\bar{v}$ by finding the zero of $H(v_x) = \kappa \, v_x^2 - \Gamma_y^\pm(v_x)$ (line~\ref{alg:velsat_solve}). If no root is found, $\bar{v}$ is set to the top speed $v_{\max}$ (line~\ref{alg:velsat_nan}). The function stores $\bar{v}$ in the vector $\boldsymbol{v}_{\mathrm{sat}}$ (line~\ref{alg:velsat_store}), which is then used in the forward and backward passes to compute the final speed and acceleration profiles.
\begin{algorithm}[]
  \caption{\textsc{VelSat} function}
  \label{alg_Vmax}
  \begin{algorithmic}[1]
    \State \textbf{Input:} $\boldsymbol s$, $\boldsymbol \kappa$, $v_{\max}$, $\Gamma_y^-(v_x)$, $\Gamma_y^+(v_x)$
    \State \textbf{Output:} $\boldsymbol{v}_{\mathrm{sat}}$
    \State $N \gets \mathrm{size}(\boldsymbol s)$\,;
    \ForAll{$i \in [1, N]$} \label{alg:velsat_for}
      \State $\kappa \gets \boldsymbol{\kappa}(i)$\,; $v_x \gets \boldsymbol{v}_x(i)$\,;
        \State \IfAlg{} $\kappa \geq 0$ \ThenAlg{} $H(V) := \kappa \, V^2 - \Gamma_y^+(V)$\,;
        \State \ElseAlg{} $H(V) := \kappa \, V^2 - \Gamma_y^-(V)$\,;
        \State $\bar{v}$ $\gets$ $\Call{solve}{H(v_x) = 0, \, [0, v_{\max}]}$\,; \label{alg:velsat_solve}
      \State \IfAlg{} $\bar{v} = \varnothing$ \ThenAlg{} $\bar{v}$ $\gets v_{\max}$\,; \label{alg:velsat_nan}
      \State $\boldsymbol{v}_{\mathrm{sat}}(i) \gets \bar{v}$\,; \label{alg:velsat_store}
    \EndFor
  \end{algorithmic}
\end{algorithm}
\subsubsection{Forward Pass}
The middle plot of Fig. \ref{fig:algorithm_phases} shows the forward pass (\textsc{F}) in Algorithm \ref{alg_FW}, which computes the maximum feasible longitudinal accelerations satisfying the g-g-v constraints \eqref{eq_OCP_ay_constr}-\eqref{eq_OCP_ax_constr}. For each path segment, the algorithm extracts the initial/final curvatures $\{\kappa_0,\kappa_1\}$ and the initial velocity $v_0$ (line~\ref{alg:fw_kappa}). It then computes the acceleration bounds $\{a_{x_{\min}}, a_{x_{\max}}\}$ (line~\ref{alg:fw_ax_bounds}) and checks whether applying $a_{x_0} = a_{x_{\max}}$ satisfies the g-g-v constraints at the segment end (lines~\ref{alg:fw_G}-\ref{alg:fw_if_max}). If not, it solves a root-finding problem to find the maximum feasible $a_{x_0}$ that brings the end state to the g-g-v boundary (line~\ref{alg:fw_if_min_solve}). The final velocity $v_1$ is then set to the minimum of $\boldsymbol{v}_{\mathrm{sat}}$ and the velocity from forward integration of $a_{x_0}$ via \eqref{eq_vx_s_no_constr} (line~\ref{alg:fw_v1_end}). This $v_1$ becomes the initial velocity for the next segment (line~\ref{alg:fw_v0_next}). If instead no feasible $a_{x_0}$ is found at line~\ref{alg:fw_if_min_solve}, $a_{x_0}$ is set to $\varnothing$, and the next segment starts from the saturated velocity $\boldsymbol{v}_{\mathrm{sat}}$ (line~\ref{alg:fw_if_min_nan}). The function finally returns the vectors of velocity $\boldsymbol{v}_x$ and acceleration $\boldsymbol{a}_x$.
\begin{algorithm}[]
  \caption{\textsc{Forward} (\textsc{F}) function}
  \label{alg_FW}
  \begin{algorithmic}[1]
    \State \textbf{Input:} $\boldsymbol s$, $\boldsymbol \kappa$, $v_{\mathrm{ini}}$, $\Gamma_y^-$, $\Gamma_y^+$, $\Gamma_x^-$, $\Gamma_x^+$, $\boldsymbol{v}_{\mathrm{sat}}$
    \State \textbf{Output:} $\boldsymbol{v}_{x}, \boldsymbol{a}_{x}$
    \State $\boldsymbol{v}_{x}, \, \boldsymbol{a}_{x} \gets \varnothing$\,;
    \State $\boldsymbol{v}_{x}(1) \gets v_{\mathrm{ini}}$ ; $v_0 \gets v_{\mathrm{ini}}$ ; $N \gets \mathrm{size}(\boldsymbol s)$\,;
    \ForAll{$i \in [1, N-1]$}
      \State $\kappa_0 \gets \boldsymbol{\kappa}(i)$ ; $\kappa_{1} \gets \boldsymbol{\kappa}(i+1)$ ; $L \gets \boldsymbol{s}(i+1) - \boldsymbol{s}(i)$\,; \label{alg:fw_kappa}
      \State $a_{y_{\mathrm{clip}}} \gets \Call{clip}{\kappa_0 \, v_0^2, \; \Gamma_y^-(v_0), \; \Gamma_y^+(v_0)}$\,;
      \State $a_{x_{\mathrm{min}}} \gets \Gamma^-(a_{y_{\mathrm{clip}}}, v_0)$\,; \quad $a_{x_{\mathrm{max}}} \gets \Gamma^+(a_{y_{\mathrm{clip}}}, v_0)$\,; \label{alg:fw_ax_bounds}
      \State $ G(A) := \textsc{D}^\pm\biggl(A, \, 
      \underbrace{\vphantom{\sqrt{v_0^2}} \kappa_1 \big(2 \, L \, A + v_0^2\big)}_{a_{y_1}}, \, 
      \underbrace{\big(2 \, L \, A + v_0^2\big)^{1/2}}_{v_1}\biggr)$\,; \label{alg:fw_G}
      \State \IfAlg{} {$G(a_{x_{\max}}) \leq 0$} \ThenAlg{} $a_{x_0} \gets a_{x_{\mathrm{max}}}$\,; \label{alg:fw_if_max}
      \State \ElseAlg{} $a_{x_0} \gets \Call{solve}{G(a_{x_0}) = 0, [a_{x_{\mathrm{min}}}, a_{x_{\mathrm{max}}}]}$\,; \label{alg:fw_if_min_solve}
      \State \IfAlg{} $a_{x_0} = \varnothing$ \ThenAlg{} $v_1 \gets \boldsymbol{v}_{\mathrm{sat}}(i+1)$\,; \label{alg:fw_if_min_nan}
      \State \ElseAlg{} $v_1 \gets \min\left(\sqrt{2 \, L \, a_{x_0} + v_0^2}, \, \boldsymbol{v}_{\mathrm{sat}}(i+1) \right)$\,;\label{alg:fw_v1_end}
      \State $\boldsymbol{a}_{x}(i) \gets a_{x_0}$ ; \, $\boldsymbol{v}_{x}(i+1) \gets v_1$ ; \, $v_0 \gets v_1$\,; \label{alg:fw_v0_next}
    \EndFor
  \end{algorithmic}
\end{algorithm}
\subsubsection{Backward Pass}
\begin{algorithm}[]
  \caption{\textsc{Backward} (\textsc{B}) function}
  \label{alg_BW}
  \begin{algorithmic}[1]
    \State \textbf{Input:} $\boldsymbol s$, $\boldsymbol \kappa$, $\Gamma_y^-$, $\Gamma_y^+$, $\Gamma_x^-$, $\Gamma_x^+$, $\boldsymbol{v}_{\mathrm{sat}}$, $\boldsymbol{v}_{x}$, $\boldsymbol{a}_{x}$
    \State \textbf{Output:} $\boldsymbol{v}_{x}$, $\boldsymbol{a}_{x}$
    \State $N \gets \mathrm{size}(\boldsymbol s)$\,;
    \ForAll{$i \in [N,2]$} \label{alg:bw_for}
      \State $\kappa_0 \gets \boldsymbol \kappa(i-1)$ ; $\kappa_1 \gets \boldsymbol \kappa(i)$ ; $v_0 \gets \boldsymbol{v}_{x}(i-1)$ ; $v_1 \gets \boldsymbol{v}_{x}(i)$\,;
      \State $a_{x_0} \gets \boldsymbol{a}_{x}(i-1)$ ; $L \gets \boldsymbol s(i) - \boldsymbol s(i-1)$\,;
      \State $\mathrm{is\_valid\_forward} \gets a_{x_0} \neq \varnothing \;\; \textbf{and} \;\; 
      \left(v_0^2 + 2 \, L \, a_{x_0}\right)^{1/2} = v_1$\,; \label{alg:bw_is_valid_forward}
      \State \IfAlg{} $\mathrm{is\_valid\_forward}$ \ThenAlg \, \ContinueAlg{} \label{alg:bw_continue}
      \State $a_{y_{\mathrm{clip}}} \gets \Call{clip}{\kappa_1 \, v_1^2, \, \Gamma_y^-(v_1), \, \Gamma_y^+(v_1)}$\,;
      \State $a_{x_{\mathrm{min}}} \gets \Gamma^-(a_{y_{\mathrm{clip}}}, v_1)$\,; \quad $a_{x_{\mathrm{max}}} \gets \Gamma^+(a_{y_{\mathrm{clip}}}, v_1)$\,; \label{alg:bw_ax_bounds}
      \State $a_{x_{\mathrm{avg}}} \gets (v_1^2 - v_0^2) / (2 \, L)$\,; \label{alg:bw_avg_acc}
      \State $v_{0_{\textsc{reach}_{\max}}} \gets \left(v_1^2-2 \, L \, a_{x_{\mathrm{min}}}\right)^{1/2}$\,;
      \State $v_{0_{\textsc{reach}_{\min}}} \gets \left(v_1^2-2 \, L \, a_{x_{\mathrm{max}}}\right)^{1/2}$\,;
      \State $\mathrm{is\_valid\_}v_0 \gets (v_0 \leq v_{0_{\textsc{reach}_{\max}}}) \; \textbf{and} \; (v_0 \geq v_{0_{\textsc{reach}_{\min}}})$\,; \label{alg:bw_vx_bounds}
      \State $\mathrm{is\_valid\_}\,a_x \gets (a_{x_{\mathrm{avg}}} \leq a_{x_{\mathrm{max}}}) \; \textbf{and} \; (a_{x_{\mathrm{avg}}} \geq a_{x_{\mathrm{min}}})$\,; \label{alg:bw_avg_acc_bounds}
      \State $G(A) := \textsc{D}^\pm\biggl(A,  
        \underbrace{\vphantom{\sqrt{v_1^2}}\kappa_0 (v_1^2-2 \, L \, A)}_{a_{y_0}},
        \underbrace{\left(v_1^2-2 \, L \, A\right)^{1/2}}_{v_0}\biggr)$ \label{alg:bw_G}
      \If{$\mathrm{is\_valid\_}\,a_x \; \textbf{and} \; \mathrm{is\_valid\_}v_0 \; \textbf{and} \;  G(a_{x_{\mathrm{avg}}}) \leq 0$} \label{alg:bw_if_avg_acc}
        \State $\boldsymbol{a}_{x}(i-1) \gets a_{x_{\mathrm{avg}}}$\,; \label{alg:bw_acc_avg_case}
      \Else \label{alg:bw_else}
        \State \IfAlg{} {$G(a_{x_{\min}}) \leq 0$} \ThenAlg{} $a_{x_0} \gets a_{x_{\mathrm{min}}}$\,; \label{alg:bw_if_min}
        \State \ElseAlg{} $a_{x_0} \gets \Call{solve}{G(a_{x_0}) = 0, [a_{x_{\mathrm{min}}}, a_{x_{\mathrm{max}}}]}$\,; \label{alg:bw_if_max_solve}
        \State $\boldsymbol{v}_{x}(i-1) \gets \left(v_1^2-2 \, L \, a_{x_0}\right)^{1/2}$\,;\; $\boldsymbol{a}_{x}(i-1) \gets a_{x_0}$\,; \label{alg:bw_update} 
      \EndIf
    \EndFor
  \end{algorithmic}
\end{algorithm}
The backward pass, described in Algorithm \ref{alg_BW} and Fig. \ref{fig:algorithm_phases} (bottom plot), computes the braking manoeuvers for all path segments where the forward pass did not yield valid solutions.
It returns the final velocity and acceleration vectors, $\boldsymbol{v}_{x}$ and $\boldsymbol{a}_{x}$. The algorithm iterates through the segments in reverse order (starting from the end, line~\ref{alg:bw_for}) and proceeds according to the following three cases.

In the first case, if the forward solution is still valid—\ie{} the final segment velocity $v_1$ is reachable from the initial state $v_0$ using the acceleration $a_{x_0}$—the algorithm proceeds to the next segment without modification (lines~\ref{alg:bw_is_valid_forward}-\ref{alg:bw_continue}).

In the second case, the initial velocity $v_0$ is still reachable from the updated final speed $v_1$, within the acceleration bounds (line~\ref{alg:bw_ax_bounds}). Here, the algorithm computes the average acceleration $a_{x_{\mathrm{avg}}}$ (line~\ref{alg:bw_avg_acc}) and checks whether it lies within the g-g-v envelope at both segment endpoints (lines~\ref{alg:bw_avg_acc_bounds}-\ref{alg:bw_if_avg_acc}). If so, $a_{x_{\mathrm{avg}}}$ is accepted (line~\ref{alg:bw_acc_avg_case}), and the algorithm continues to the next segment.

In the final case, either $a_{x_{\mathrm{avg}}}$ is infeasible or $v_0$ is no longer reachable from $v_1$ (line~\ref{alg:bw_else}). The algorithm then applies the maximum deceleration $a_{x_{\min}}$, if it remains within the g-g-v envelope (line~\ref{alg:bw_if_min}), or computes the deceleration $a_{x_0}$ that brings $v_0$ to the g-g-v envelope boundary (line~\ref{alg:bw_if_max_solve}).
Finally, the updated velocity and acceleration are stored (line~\ref{alg:bw_update}).
The backward pass always yields a feasible solution, and overrides the requested initial velocity $v_{\mathrm{ini}}$ if it is infeasible.
%
%
\section{RESULTS} \label{sec:results}
\subsection{Simulation Setup, Implementation and Benchmarking}
We validate the proposed FBGA with the following simulation setup, implementation code, and benchmarking.
\subsubsection{Vehicle Types and Racetracks} \label{sec:vehicle_racetrack}
FBGA could be applied for velocity planning with different robotic systems (Sec. \ref{sec:intro}), and in this paper, we focus on the challenging case of racing vehicles, whose g-g-v acceleration envelopes can have complex non-convex shapes. We consider two vehicle types: 
a racing car, whose g-g-v envelope is taken from \cite{piccinini2025howoptimal}, and 
a racing motorcycle, whose g-g-v is derived from \cite{Biral2009}. 
For each vehicle type, we apply FBGA on five racetracks: Catalunya, Valencia, Misano, Sepang, and Palm Beach.
\subsubsection{Implementation}
Our code is implemented in \texttt{C++} and is available open source at \url{https://github.com/DRIVEWISE/FBGA}. 
Our numerical experiments are run on a laptop with an M2 Max Apple Silicon chip.
\subsubsection{Benchmarking}
We benchmark FBGA against OCP$_{\mathrm{bench}}$, which is a numerical solution of the same fixed-path problem \eqref{eq_OCP_problem}, obtained using the state-of-the-art indirect optimal control solver Pins \cite{biral2016notes}.
\subsection{Preprocessing: Path Generation} \label{sec:preprocess}
Our FBGA computes the time-optimal speed profile along a given path.
For each racetrack and vehicle (Sec. \ref{sec:vehicle_racetrack}), the path is generated by solving a free-trajectory minimum-lap-time OCP (MLT-OCP), using the formulation in \cite{piccinini2025howoptimal} and the solver Pins \cite{biral2016notes}. The MLT-OCP uses a point-mass model with longitudinal and lateral dynamics, and enforces the same g-g-v constraints as in \eqref{eq_OCP_problem}. From the resulting minimum-time lap, we extract the curvilinear abscissas $\boldsymbol s$ and curvatures $\boldsymbol \kappa$, used for both our FBGA and OCP$_\mathrm{bench}$. We remark that the path can be obtained with other methods according to the application (\eg{} from experimental data, or geometrical optimization), and does not need to be generated by an OCP.
\subsection{Maneuver Analysis} \label{sec:results_maneuver}
\begin{figure}[]
  \centering
  \includegraphics[width=0.98\columnwidth]{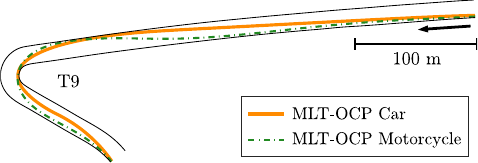}
  \caption{Comparison of the MLT-OCP maneuvers for the racing car and motorcycle models of this paper, on the turn n.9 of the Sepang circuit. 
  }
  \label{fig:MLTvsFWBW_TRAJ}
  \vspace{-0.1cm}
\end{figure}
\begin{figure}[]
  \centering
  \subfloat[Results for the racing car model.]{\includegraphics[width=0.98\columnwidth]{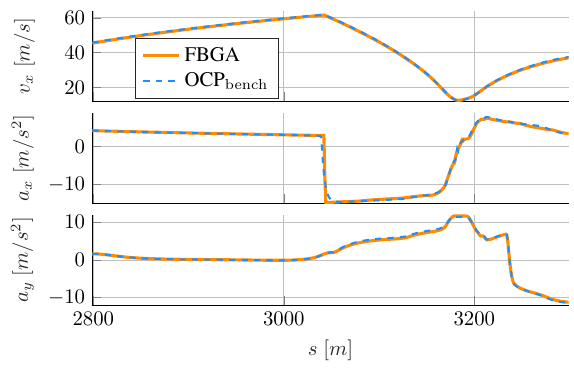 }\label{fig:MLTvsFWBW_CAR}}\\
  \subfloat[Results for the racing motorcycle model.]{\includegraphics[width=0.98\columnwidth]{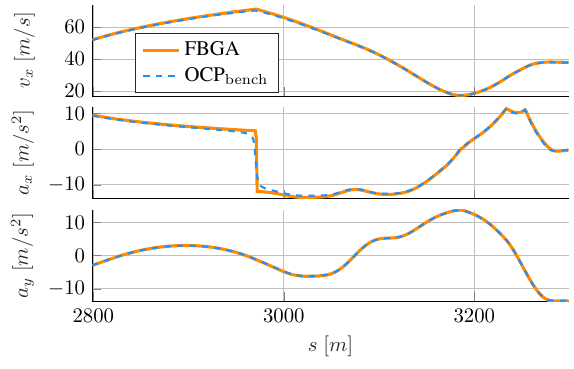}\label{fig:MLTvsFWBW_MOTO}}
  \caption{Velocity, longitudinal and lateral acceleration profiles of our FBGA and the benchmark OCP$_{\mathrm{bench}}$, for a racing car (a) and motorcycle (b) at turn 9 of the Sepang circuit (Fig.~\ref{fig:MLTvsFWBW_TRAJ}).
  }
  \label{fig:MLTvsFWBW}
  \vspace{-0.1cm}
\end{figure}
\begin{figure}[]
  \centering
  \subfloat[Top view]{\includegraphics[width=0.35\columnwidth]{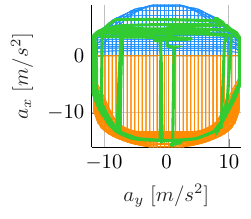}\label{fig:gg_diagram_and_ctrl_trajFront}}
  \subfloat[Lateral view]{\includegraphics[width=0.60\columnwidth]{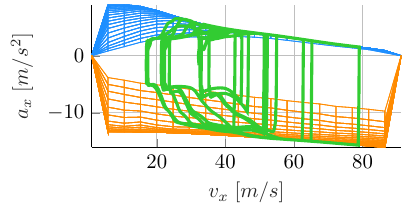 }\label{fig:gg_diagram_and_ctrl_trajLAT}}
  \caption{g-g-v envelope (blue-orange lines) for the race car model, and solution computed by our FBGA (green line) on the Catalunya circuit.}
  \label{fig:gg_diagram_and_ctrl_traj}
  \vspace{-0.1cm}
\end{figure}

We now compare the results of our FBGA with the benchmark OCP$_{\mathrm{bench}}$, on the five circuits and the race car and motorcycle models described in Sec. \ref{sec:vehicle_racetrack}. The two vehicles have markedly different g-g-v acceleration constraints, shown in Fig. \ref{fig:ggvA} (motorcycle) and Fig. \ref{fig:gg_diagram_and_ctrl_traj} (car). The motorcycle's envelope is notably non-convex due to the front and rear wheel lift during acceleration and braking, which limits the peak longitudinal accelerations \cite{Biral2009,piccinini2024impacts}.

Fig. \ref{fig:MLTvsFWBW_TRAJ}-\ref{fig:MLTvsFWBW} shows the minimum-time maneuvers (path, speed, and accelerations) at turn 9 of the Sepang circuit, for both the car and motorcycle models. The paths in Fig. \ref{fig:MLTvsFWBW_TRAJ} are computed via the MLT-OCP preprocessing step, which accounts for the g-g-v constraints and vehicle transient dynamics (Sec. \ref{sec:preprocess}). Due to their distinct g-g-v envelopes, the resulting time-optimal paths differ significantly: the car brakes on a straight before the turn, while the motorcycle brakes along a curved path to reach higher longitudinal accelerations by avoiding rear-wheel lift. 
Fig. \ref{fig:MLTvsFWBW} plots the corresponding speed and acceleration profiles. The braking and acceleration phases vary notably between the car (Fig. \ref{fig:MLTvsFWBW_CAR}) and motorcycle (Fig. \ref{fig:MLTvsFWBW_MOTO}), again reflecting their differences in the g-g-v diagrams. FBGA's results closely match those of OCP$_\mathrm{bench}$, with minor discrepancies in transition regions where FBGA switches instantaneously between acceleration and braking. Conversely, OCP$_\mathrm{bench}$ filters the control $a_x$ with a fast first-order dynamics, to prevent numerical oscillations of the solution around the g-g-v boundaries. Fig. \ref{fig:gg_diagram_and_ctrl_trajFront}-\ref{fig:gg_diagram_and_ctrl_trajLAT} plots the g-g-v diagram of the racing car model, together with the solution computed by FBGA (green line) on the Catalunya circuit. The figures highlight that the FBGA solution moves along the g-g-v envelope, while satisfying the constraints up to the desired solver's precision. 
\begin{figure}[]
  \centering
  \includegraphics[width=0.98\columnwidth]{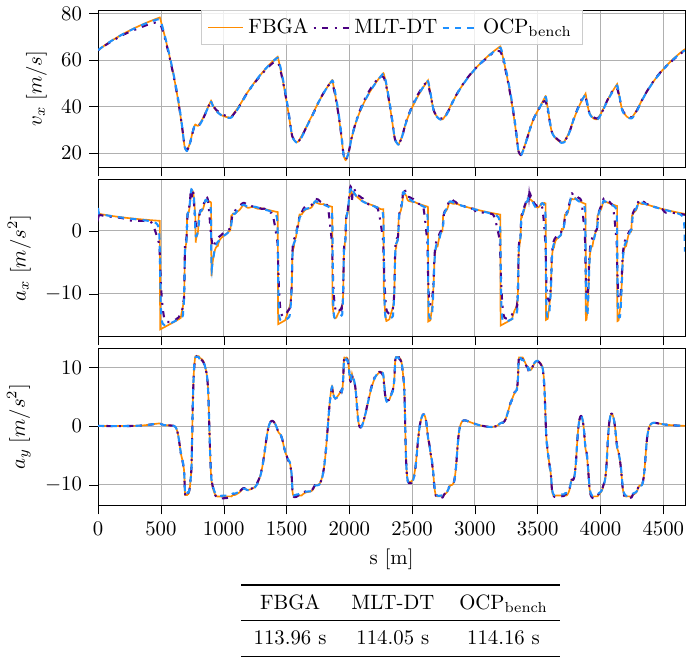}
  \vspace{-0.1cm}
  \caption{Comparing our FBGA, the OCP$_{\mathrm{bench}}$ and the MLT-DT OCP, which is solved with a double-track (DT) vehicle model, on the Catalunya circuit.}
  \label{fig:MLTvsFWBWvsOCP}
  \vspace{-0.1cm}
\end{figure}
To further validate our FBGA, we compare it with a minimum-lap-time OCP solved using a high-fidelity double-track vehicle model (MLT-DT), formulated as in \cite{PagotPhDThesis}. This model captures high-speed race car dynamics, including a Pacejka MF5.2 tire model, real engine torque curves, and a limited-slip differential. We apply FBGA on the path generated by MLT-DT on the Catalunya circuit, and the g-g-v envelope is identified following \cite{piccinini2025howoptimal}. As shown in Fig.~\ref{fig:MLTvsFWBWvsOCP}, our approach closely matches both MLT-DT and OCP$_\mathrm{bench}$, with minor deviations only during acceleration-braking transitions, where jerk is limited in the double-track model. Despite this, the lap time difference between FBGA and MLT-DT is only $73\;\mathrm{ms}$, relatively small given the model complexity and the different optimization methods.
\subsection{Lap Times} \label{sec:lap_times}
\begin{table*}[]
	\centering
	\begin{tabular}{c|c|c|c|c|c|c|c}
		\toprule 
    \multirow{2}{*}{\textbf{Circuit (length)}} & \textbf{N. mesh} & \multirow{2}{*}{\textbf{Vehicle}} &  \multicolumn{3}{c|}{\textbf{Lap time [s]}} & \multicolumn{2}{c}{\textbf{CPU time [ms]}} \\ 
                                          & \textbf{points}         &                  & OCP$_\mathrm{bench}$       & \textbf{FBGA} (ours) & $\Delta$ (FBGA$\,-\,$OCP$_\mathrm{bench}$) & OCP$_\mathrm{bench}$ & \textbf{FBGA} (ours) \\
    \cline{1-8}
    \multirow{2}{*}{Catalunya (4.66 km)}  & \multirow{2}{*}{4660}   & Car              & $112.461$ & $112.204$ & $-0.257$ & $8017.15$ &  $9.86$ \\
                                          &                         & Motorcycle       & $105.381$ & $104.999$ & $-0.382$ &  $693.91$ &  $3.31$ \\
    \cline{1-8}
    \multirow{2}{*}{Valencia (4.00 km)}   & \multirow{2}{*}{4000}   & Car              & $104.742$ & $104.520$ & $-0.222$ & $7411.99$ &  $8.85$ \\
                                          &                         & Motorcycle       &  $96.752$ &  $96.434$ & $-0.318$ &  $830.82$ &  $3.40$ \\
    \cline{1-8}
    \multirow{2}{*}{Misano (4.16 km)}     & \multirow{2}{*}{4160}   & Car              & $107.740$ & $107.451$ & $-0.289$ & $6805.87$ &  $9.66$ \\
                                          &                         & Motorcycle       & $98.814$ &  $98.497$ & $-0.317$ &  $602.19$ &  $3.37$ \\
    \cline{1-8}
    \multirow{2}{*}{Sepang (5.52 km)}     & \multirow{2}{*}{5520}   & Car              & $135.480$ & $135.086$ & $-0.394$ & $8239.47$ & $11.44$ \\
                                          &                         & Motorcycle       & $125.483$ & $125.034$ & $-0.449$ & $1282.58$ &  $4.30$ \\
    \cline{1-8}
    \multirow{2}{*}{Palm Beach (3.17 km)} & \multirow{2}{*}{3170}   & Car              & $79.963$ & $79.869$ & $-0.094$ & $4819.43$ &  $6.38$ \\
                                          &                         & Motorcycle       & $74.790$ & $74.537$ & $-0.253$ &  $706.43$ &  $2.39$ \\
		\toprule
	\end{tabular}
	\caption{Comparison of lap times and CPU times between our FBGA and the benchmark OCP$_\mathrm{bench}$, both solved with the same mesh, on an M2 Max Apple Silicon chip. The table shows the results for the racing car and motorcycle models on $5$ racetracks.}
	\label{tab_full_lap}
  \vspace{-0.3cm}
\end{table*}
Table~\ref{tab_full_lap} compares the lap times of FBGA and OCP$_\mathrm{bench}$ for the race car and motorcycle models, across five racetracks. The lap times of FBGA and OCP$_\mathrm{bench}$ are very close, with differences ranging from $0.094$ to $0.449$\,s ($0.11\%-0.36\%$ of the total lap time). These deviations are comparable to those observed among professional race drivers~\cite{Kegelman2017}.
FBGA slightly underestimates the lap times due to instantaneous traction-braking transitions, while professional drivers have a limited actuation rate.
Nonetheless, this simplification boosts the computational efficiency while still providing accurate lap time estimates. Indeed, FBGA is designed for high-level trajectory planning \cite{Ogretmen2024,Romero2022}, where the computational time is key, and the acceleration rate bounds are typically handled by downstream motion controllers.
\subsection{Computational Times}
\subsubsection{Full-Lap Computational Times}
Table \ref{tab_full_lap} reports the computational (CPU) times of the FBGA and OCP$_\mathrm{bench}$ methods when computing the speed profiles over full laps. The CPU times of FBGA range in $2.39-11.44$ ms, and they are $2-3$ orders of magnitude lower\footnote{As a remark, the CPU times are obtained with single runs of our FBGA, without averaging over multiple runs.} than OCP$_\mathrm{bench}$. The CPU times of both methods are influenced by the path curvature values\footnote{A CPU time analysis of our FBGA shows that roughly $20\%$ of the time is spent in the \textsc{VelSat} function, $60\%$ in the \textsc{Forward} and $20\%$ in the \textsc{Backward} functions. These ratios highly depend on the provided path.} and the implementation of the g-g-v constraints \eqref{eq_OCP_ay_constr}-\eqref{eq_OCP_ax_constr}. For the car, these constraints are modeled with bilinear splines, while for the motorcycle, we use analytical functions, which are computationally cheaper.\\
These results show that our FBGA algorithm supports any formulation of the g-g-v acceleration constraints, independently of their complexity, without the need for convexity or differentiability assumptions.
\subsubsection{Short-Horizon Computational Times}
We evaluate the performance of our FBGA for short-horizon planning over a $300$ m segment, covering the first two corners of the Catalunya circuit (Fig. \ref{fig:3DTraj}). This horizon length is typically adopted in sampling-based planners for high-speed autonomous racing \cite{Ogretmen2024,piazza2024mptree}.
Table \ref{tab_short_horizon} reports the results for different mesh sizes used to discretize the path. It compares our FBGA with OCP$_\mathrm{bench}$, which uses a fixed number of $300$ mesh points. With the same mesh, FBGA is computationally faster by a factor of $550$ ($0.177$ ms vs $97.533$ ms), with only a $0.19\%$ difference in maneuver time ($8.7765$ s vs $8.7936$ s). Reducing the FBGA's mesh size to $100$ points further increases the speed-up to $3$ orders of magnitude, with negligible impact on the maneuver time. 
These analyses show that our FBGA is robust to the mesh resolution and suitable for real-time planning of multiple time-optimal maneuvers, as required in sampling-based planners for autonomous racing \cite{Ogretmen2024}. 
\begin{table}[]
	\centering
	\begin{tabular}{c|c||c|c|c}
		\toprule 
    \multicolumn{2}{c||}{\textbf{OCP$_\mathrm{bench}$}} & \multicolumn{3}{c}{\textbf{FBGA} (ours)} \\ 
    \textbf{T [s]} & \textbf{CPU [ms]} & \textbf{N. points} & \textbf{T [s]} & \textbf{CPU [ms]} \\
    \cline{1-5}
    \multirow{3}{*}{8.7936} & \multirow{3}{*}{97.533} & 300 & 8.7765 & 0.177 \\
     &  & 200 & 8.7830 & 0.131 \\
     &  & 100 & 8.8029 & 0.062 \\
		\toprule
	\end{tabular}
	\caption{Comparing the maneuver time $T$ and CPU times of our FBGA and OCP$_\mathrm{bench}$, on a $300$ m horizon. OCP$_\mathrm{bench}$ uses 300 mesh points, while FBGA is solved with a variable mesh.}
	\label{tab_short_horizon}
  \vspace{-0.4cm}
\end{table}
\subsection{Sensitivity Analysis}
\begin{figure}[htb]
  \centering
  \includegraphics[width=0.99\columnwidth]{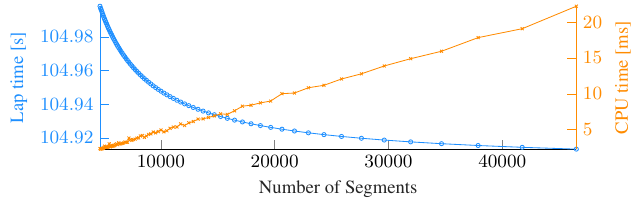}
  \vspace{-0.1cm}
  \caption{Lap times and CPU times of FBGA as a function of the number of mesh segments, with the racing motorcycle model on the Catalunya circuit.
  }
  \label{fig:sensitivity_cpp_catalunya}
  \vspace{-0.1cm}
\end{figure}

Fig. \ref{fig:sensitivity_cpp_catalunya} shows the sensitivity of our FBGA algorithm to the number of path discretization segments, analyzing both the lap time (Catalunya circuit) and the total CPU time. As expected, the CPU time scales linearly with the number of segments\footnote{Our results are obtained by randomizing the number of segments at each run, to avoid caching optimization of subsequent calls of the same functions.}, since the algorithm processes each segment sequentially. Minor oscillations appear at coarse resolutions, where segment-level curvature variations can affect the number of iterations of our root-finding procedure. In contrast, the lap time decreases hyperbolically, and changes by only 0.07 s (0.06\%) when increasing the segment count by 10 times. \\
These results confirm that FBGA is robust to the mesh resolution, and can be applied to different use cases: fine meshes maximize the solution accuracy and fully exploit the acceleration limits, while coarse meshes offer significant speed-ups with a negligible impact on the maneuver time.
%
%
\section{CONCLUSIONS}
We introduced FBGA, a new forward-backward method for real-time computation of time-optimal velocity profiles along prescribed paths, supporting arbitrary acceleration constraints. FBGA discretizes the path into short segments and performs forward and backward integrations to maximize the velocity while satisfying user-defined acceleration limits.\\
FBGA was validated in the context of autonomous racing, with race car and motorcycle models on five circuits. It produced velocity and acceleration profiles closely matching those of a benchmark optimal control problem ($\mathrm{OCP}_{\mathrm{bench}}$), with lap time deviations below $0.449$ s ($0.36\%$) and computational speed-ups of up to three orders of magnitude. 
These results held across tracks and vehicle types, with no assumptions on convexity, smoothness, or differentiability of the acceleration constraints.
In short-horizon planning, FBGA achieved maneuver times within $0.19\%$ of $\mathrm{OCP}_{\mathrm{bench}}$, while maintaining a three orders of magnitude speed-up and a high solution quality even with coarse path discretization. \\
Overall, FBGA can be applied to accurate full-lap velocity optimization, warm-starting of complex \acp{OCP}, and real-time short-horizon planning. It is expected to be an effective building block for multi-query time-optimal trajectory planning in dynamic environments, extending sampling-based planners \cite{piazza2024mptree,Romero2022} beyond conservative box acceleration constraints, and unlocking new capabilities for autonomous racing.\\
Future work will enhance the FBGA method to address its limitations and extend its scope. We plan to improve the feasibility of the acceleration-braking transitions by extending Algorithms \ref{alg_FW}-\ref{alg_BW} to deal with jerk limits. Also, formal guarantees on the solution's optimality and convergence will be better investigated. Finally, application-oriented papers will integrate FBGA into multi-query trajectory planners for diverse fields of robotics, including vehicles and drone flight.
%
%
%
\bibliographystyle{IEEEtran}
\bibliography{ieeetran.bib}
\end{document}